\documentclass[runningheads]{llncs}
\usepackage{amsmath,amsfonts}
\usepackage{algorithmic}
\usepackage{algorithm}
\usepackage{array}
\usepackage[caption=false,font=normalsize,labelfont=sf,textfont=sf]{subfig}
\usepackage{textcomp}
\usepackage{stfloats}
\usepackage{url}
\usepackage{verbatim}
\usepackage{graphicx}
\usepackage{cite}
\usepackage{color}
\usepackage{microtype}
\usepackage{multirow}
\usepackage{amssymb}
\usepackage{adjustbox}
\usepackage{makecell}
\usepackage{arydshln}
\usepackage{hyperref}
\usepackage{tablefootnote}
\usepackage[inline]{enumitem}
\usepackage[normalem]{ulem}
\usepackage[dvipsnames]{xcolor}
\usepackage{orcidlink}
\usepackage{soul}
\usepackage{wrapfig}

\usepackage{pifont}

\DeclareMathOperator{\KD}{KD}

\DeclareMathOperator{\KL}{KL}

\DeclareMathOperator{\NLL}{NLL}

\DeclareMathOperator{\VAE}{VAE}
\DeclareMathOperator{\softmax}{\texttt{softmax}}
\DeclareMathOperator{\diag}{\texttt{diag}}

\begin{document}

\title{Improving Neural Topic Models with Wasserstein Knowledge Distillation}

\author{Suman Adhya\,\orcidlink{0000-0001-6568-2020} \and Debarshi Kumar Sanyal\,\orcidlink{0000-0001-8723-5002}}

\authorrunning{S. Adhya and D. K. Sanyal}
\institute{Indian Association for the Cultivation of Science, Jadavpur 700032, India
\email{adhyasuman30@gmail.com, debarshisanyal@gmail.com }}

\maketitle

\begin{abstract}
Topic modeling is a dominant method for exploring document collections on the web and in digital libraries. Recent approaches to topic modeling use pretrained contextualized language models and variational autoencoders. However, large neural topic models have a considerable memory footprint. In this paper, we propose a  knowledge distillation framework to compress a contextualized topic model without loss in topic quality. In particular, the proposed distillation objective is to minimize the cross-entropy of the soft labels produced by the teacher and the student models, as well as to minimize the squared 2-Wasserstein distance between the latent distributions learned by the two models. Experiments on two publicly available datasets show that the student trained with knowledge distillation achieves topic coherence much higher than that of the original student model, and even surpasses the teacher while containing far fewer parameters than the teacher's. The distilled model also outperforms several other competitive topic models on topic coherence.
\end{abstract}
\section{Introduction}
\label{sec:intro}
Topic modeling has come up as an important technique to analyze large document corpora and extract their themes automatically \cite{adhya-sanyal-2022-indian}, \cite{HTKG},  \cite{vayansky2020review}. Therefore, they are frequently used to obtain an overview of the topics in document archives and web search results, match queries and documents, and diversify search results \cite{zhai2018tutorial, guo2022semantic}.
While latent Dirichlet allocation (LDA) \cite{blei2003latent} is the classical topic modeling algorithm, recent approaches exploit deep neural networks, specifically, variational autoencoders (VAEs) \cite{kingma2013auto}. ProdLDA \cite{srivastava2017autoencoding} is a well-known VAE-based topic model that uses a product of experts and a Laplace approximation to the Dirichlet prior.  
Bianchi et al. \cite{bianchi2021pre} recently proposed \textit{CombinedTM}, a contextualized topic model that feeds into the VAE of ProdLDA a distributed representation of the document built with a pre-trained language model (PLM) like sentence-BERT (SBERT) \cite{reimers2019sentence} along with a bag-of-words (BoW) representation of the document. It achieves state-of-the-art topic coherence on many benchmark data sets. Given a VAE-based topic model pre-trained on a corpus, one can pass a document from the corpus through the VAE encoder and recover its topics.  
A remarkable feature of contextualized topic models is that, if the PLM is multilingual and the input to the encoder solely consists of contextualized representations from the PLM, it is possible to train the model in one language and test it in another, making it a zero-shot topic model, also called \textit{ZeroShotTM} \cite{bianchi2021cross}. 
Increasing the network complexity, like the depth or width of the neural networks in the VAE, might improve the coherence of the generated topics but produce a larger memory footprint, thereby making it difficult to store and use the topic models on resource-constrained devices. Using only contextualized embeddings in the input would also reduce the model size but could impact the topic quality as well. 

In this paper, we investigate if a VAE-based topic model can be compressed without compromising topic coherence. 
For this purpose, we use knowledge distillation (KD), which involves a teacher model to improve the performance of a smaller student model \cite{hinton2015distilling}. While KD has been used for classification tasks in image \cite{gou2021knowledge} and text processing \cite{nityasya2022student},  this paper tackles an unsupervised learning problem for a generative model. Specifically, we distill knowledge from a CombinedTM teacher to a smaller ZeroShotTM student. In standard KD \cite{hinton2015distilling}, the aim is to minimize the cross-entropy between the soft labels produced by the student and the teacher models along with the Kullback-Leibler (KL) divergence between their respective output  distributions. But even if the two distributions have very little dissimilarity with each other, the KL-divergence may reach a very high value, and if the two distributions are not overlapping at all, it explodes to infinity \cite{NEURIPS2019_f9209b78}.
To avoid these issues, we choose 2-Wasserstein distance \cite{olkin1982distance} instead of KL-divergence in distillation loss. 
Our distillation process minimizes the cross-entropy between the soft labels produced by the teacher and the student, \textit{and} the square of the 2-Wasserstein distance between the latent distributions learned by the two models. Wasserstein distance arises in the theory of optimal transport and measures how `close' two distributions are  \cite{panaretos2019statistical,  villani2021topics, gao2022distributionally}. Unlike the KL divergence, if the Wasserstein between two distributions is high, this actually represents that the underlying distributions are very different from each other. 

In summary, our contributions are:
\textbf{(1)} We propose a 2-Wasserstein distance-based knowledge distillation framework for neural topic models. We call our method \textit{Wasserstein knowledge distillation}. To the best of our knowledge, this is the first work on inter-VAE knowledge distillation for topic modeling. \textbf{(2)} Experiments on two public datasets show that in terms of topic coherence, the distilled model significantly outperforms the student and even scores better than the teacher. The distilled model also beats several strong baselines on topic coherence. This demonstrates the efficacy of our approach. We have made our code publicly available\footnote{\url{https://github.com/AdhyaSuman/CTMKD}}.

\section{Background on Wasserstein Distance}
\label{sec:WD}
Let $(\mathcal{X}, d)$ be a complete separable metric space with metric $d$ and equipped with a Borel $\sigma$-algebra. 
Let $\mathcal{P}(\mathcal{X})$ denote the space of all probability measures defined on $\mathcal{X}$ with finite $p$-th moment for $p \geq 1$. 
If $\mathbb{P}_1, \mathbb{P}_2 \in \mathcal{P}(\mathcal{X})$, then $\operatorname{\Pi}(\mathbb{P}_1, \mathbb{P}_2)$ is defined to be the set of measures ${\pi} \in \mathcal{P}(\mathcal{X}^2)$ having $\mathbb{P}_1$ and $\mathbb{P}_2$ as marginals. The $p^\text{th}$ Wasserstein distance between the two probability measures $\mathbb{P}_1$ and $\mathbb{P}_2$ in $\mathcal{P}(\mathcal{X})$ is defined as
\begin{align}
    W_p(\mathbb{P}_1, \mathbb{P}_2) &= \left( \inf_{{\pi} \in \operatorname{\Pi}(\mathbb{P}_1, \mathbb{P}_2)} \int_{\mathcal{X}^2} d({x},{y})^p \operatorname{d}{{\pi}}({x},{y}) \right)^{1/p}
\end{align}
$W_p(\mathbb{P}_1, \mathbb{P}_2)$ is intuitively the minimum `cost' of transforming $\mathbb{P}_1$ to $\mathbb{P}_2$ (or vice versa) \cite{villani2021topics}. 
Consider $\mathcal{X} = \mathbb{R}^n$ with $d$ as the Euclidean norm. Suppose $\mathbb{P}_1 = \mathcal{N}(\mu_1, \Sigma_1)$, and $\mathbb{P}_2 = \mathcal{N}(\mu_2, \Sigma_2)$ are normal distributions with means $\mu_1, \mu_2 \in \mathbb{R}^n$ and symmetric positive semi-definite covariance matrices $\Sigma_1, \Sigma_2 \in \mathbb{R}^{n \times n}$. From \cite{olkin1982distance}, the squared 2-Wasserstein distance between $\mathbb{P}_1$ and $\mathbb{P}_2$ is given by:
\begin{align}
  W_2(\mathbb{P}_1, \mathbb{P}_2)^2 = \lVert {\mu}_1-{\mu}_2 \rVert_{2}^{2}+ \operatorname{trace} \big({\Sigma}_1+ {\Sigma}_{2} -2( {\Sigma}_{2}^{1/2} {\Sigma}_{1} {\Sigma}_{2}^{1/2})^{1/2}\big)
  \label{eq:2WD}
\end{align}
Wasserstein distance has been used to train various machine learning models, including classifiers \cite{frogner2015learning}, Boltzmann machines \cite{montavon2016wasserstein}, and generative adversarial networks \cite{arjovsky2017wasserstein}, where it is found to be a better loss metric than KL-divergence.

\section{Proposed Framework for Knowledge Distillation}
\begin{figure}[!htbp]
    \centering
    \includegraphics[width=0.73\linewidth]{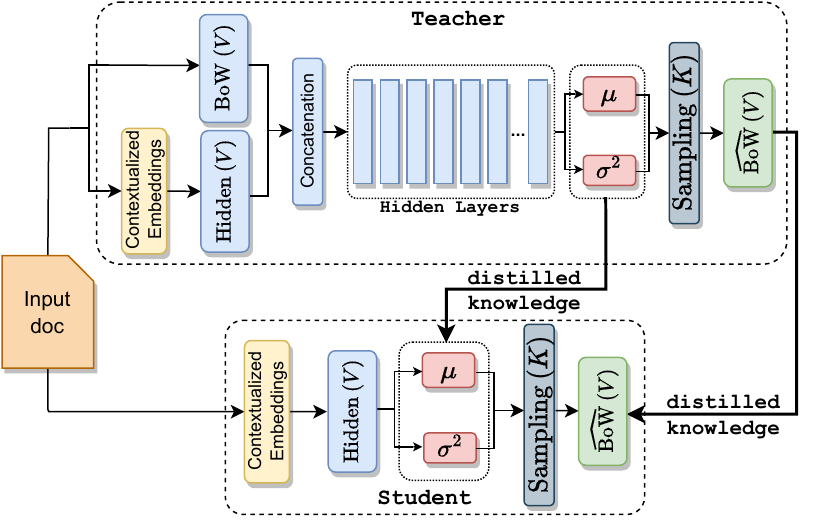}
    \caption{Framework for knowledge distillation from CombinedTM to ZeroShotTM.}
    \label{fig:KDFramework}
\end{figure}
Our framework for KD is shown in Figure \ref{fig:KDFramework}. The teacher and the student models are both VAEs.  The teacher $T$ is a CombinedTM \cite{bianchi2021pre} that takes as input ${x}$ a document encoded as the concatenation of the document's normalized BoW representation ${x}_{\text{BoW}} \in \mathbb{R}^{V}$, where $V$ is the vocabulary size, and its  contextualized embedding ${x}_{\text{ctx}}$ scaled to dimension $V$ by a linear layer. The student is a ZeroShotTM \cite{bianchi2021cross}. While the student's encoder takes only the document's contextualized representation, its decoder still needs the BoW vector during training, 
but it is not necessary when we use only its trained encoder to infer the topics for a given document. The teacher's encoder is a multi-layer feed-forward neural network (FFNN) while we make the student's encoder an FFNN with one hidden layer.

A VAE-based topic model works as follows \cite{srivastava2017autoencoding}. Suppose it has to learn $K$ topics from a corpus. The VAE encoder having weights $W$ learns the approximate posterior distribution $q_{{W}}({z} \vert {x})$  represented by mean $\mu \in \mathbb{R}^{K}$ and variance $\sigma^2 \in \mathbb{R}^{K}$ for an input instance ${x}$.  The decoder samples a vector ${z} \sim q_{W} ({z} \vert {x})$ using the reparameterization trick \cite{kingma2013auto}, and produces the document-topic vector ${\theta} = \softmax({z})$, which is passed through a shallow FFNN with weight matrix $\beta_{K \times V}$ to learn a distribution $p_{{\beta}}({x} \vert {z})$. The VAE is trained by backpropagation to minimize the following loss $\mathcal{L}_{\VAE}$:
\begin{align}
   \mathcal{L}_{\VAE} &= \mathcal{L}_{\NLL} + \mathcal{L}_{\KL}  \equiv - \mathbb{E}_{{z} \sim q_{W}({z} \vert {x})} \big[\log p_{{\beta}} ({x} \vert {z})\big] 
   + D_{\KL}\big( q_{W}({z} \vert {x}) \parallel p({z}) \big)
   \label{eq:vae_loss}
\end{align}
where $\mathcal{L}_{\NLL}$ is the expected negative log-likelihood of the reconstructed BoW, and $\mathcal{L}_{\KL}$ is a regularizer measuring the KL-divergence of the encoder's output $q_{{W}}({z} \vert {x})$ from the assumed prior $p({z})$ of the latent distribution.

Now suppose that the teacher has been already trained on a dataset to learn $K$ topics, and that, after training, the weights of its encoder and decoder are ${W}^\ast_T$ and  ${\beta}^\ast_T$, respectively. We will use this frozen teacher model to train the student \textit{with KD} to learn $K$ topics from the same dataset and the same vocabulary. We denote this KD-trained student by $S'$. Let the weights in its encoder and decoder be ${W}_{S'}$ and ${\beta}_{S'}$, respectively, at the start of some iteration during the training of $S'$. 
Given an input instance $x$, the student's loss function has two components: 

\noindent\textit{(i) Loss associated with student VAE:} The VAE loss $\mathcal{L}_{\VAE}$ is given by Eq. \eqref{eq:vae_loss}. 

\noindent\textit{(ii) Loss associated with knowledge distillation:} 
While training $S'$, every instance ${x}$ is passed through both $T$ and $S'$. Suppose the teacher's encoder outputs the $K$-variate Gaussian $\mathcal{N}({z} | {\mu}_T, {\sigma}_T^2)$ while the student's encoder outputs the $K$-variate Gaussian $\mathcal{N}({z} | {\mu}_{S'}, {\sigma}_{S'}^2)$. Note that instead of a full covariance matrix, a diagonal covariance matrix (encoded as a vector) is learned \cite{bianchi2021pre}, \cite{bianchi2021cross}. 
Let ${\Sigma}_T = \diag(\sigma^2_{T})$ and ${\Sigma}_{S'} = \diag(\sigma^2_{S'})$, which are easily observed to be  symmetric positive semi-definite. We calculate the squared 2-Wasserstein distance between the distributions learned by $T$ and $S'$ using Eq. \eqref{eq:2WD}: 
\begin{align}
   \mathcal{L}_{\text{KD-2W}} = 
   & \lVert {\mu}_{T}-{\mu}_{S'} \rVert_{2}^{2}+ \operatorname{trace} \big({\Sigma}_T+ {\Sigma}_{S'} -2( {\Sigma}_{S'}^{1/2} {\Sigma}_{T} {\Sigma}_{S'}^{1/2})^{1/2}\big)
   \label{eq:KD-2W}
\end{align}
We propose to minimize $\mathcal{L}_{\text{KD-2W}}$ so that the distribution learned by the student is pulled close to that of the teacher. 
The decoder of the teacher and that of the student produce unnormalized logits $u_T = \beta_T^\top \theta$ and $u_{S'} = \beta_{S'}^\top \theta$, respectively. We compute the cross-entropy loss $\mathcal{L}_{\text{KD-CE}}$ between the soft labels $\softmax ({u}_{T}/t)$ and $\softmax ({u}_{S'}/t)$ where $t$ is the softmax temperature (hyperparameter) \cite{hinton2015distilling}. In addition to identifying the most probable class, the soft labels formed by a higher softmax temperature ($t>1$) capture the correlation between the labels, which is desired in the distillation framework.
The total loss due to KD is 
\begin{align}
      \mathcal{L}_{\KD} = \mathcal{L}_{\text{KD-2W}} + t^2  \mathcal{L}_{\text{KD-CE}}
      \label{eq:kdloss}
\end{align}
Finally, with $\alpha \in [0,1]$ as a hyperparameter, the total loss for the student $S'$ is
\begin{align}
   \mathcal{L}_{S'} = (1-\alpha)  \mathcal{L}_{\text{VAE}} + \alpha \mathcal{L}_{\text{KD}}
\end{align}

\section{Experimental Setup}
We have performed all experiments in \href{https://github.com/MIND-Lab/OCTIS}{OCTIS}  \cite{terragni2021octis}, which is an integrated framework for topic modeling. 
We use the following datasets from OCTIS: \textbf{20NG}, which contains $16,309$ newsgroup documents on $20$ different subjects \cite{terragni2021octis}, and \textbf{M10} comprising 8355 scientific publications from 10 distinct research areas \cite{10.5555/3060832.3060886}.  
For each dataset, the vocabulary contains the 2K most common words in the corpus.
We represent each topic by its top-10 words. We use \textbf{Normalized Pointwise Mutual Information} (\textbf{NPMI}) \cite{lau-etal-2014-machine} and  \textbf{Coherence Value} (\textbf{CV}) \cite{roder2015exploring, krasnashchok2018improving} to measure topic coherence. 
NPMI of a topic is high if the words in the topic tend to co-occur. CV is calculated using an indirect cosine measure along with the NPMI score over a boolean sliding window. Higher values of NPMI and CV are better. 

The experiments are done for topic counts $K \in \{20, 50, 100\}$ on the 20NG dataset and for topic counts $K \in \{10, 20, 50, 100\}$ on the M10 dataset, where 20 and 10 are the golden number of categories for 20NG and M10, respectively. 
We denote the teacher (CombinedTM) by \textbf{T}, the student (ZeroShotTM) by \textbf{S}, and the distilled student model (ZeroShotTM)  by \textbf{SKD}. The encoder in \textbf{T} uses 768-dimensional contextualized sentence embeddings (SBERT) from   {\href{https://huggingface.co/sentence-transformers/paraphrase-distilroberta-base-v2}{\texttt{paraphrase-distilroberta-base-v2}}}. The encoders  in \textbf{S} and \textbf{SKD} use 384-dimensional SBERT embeddings from {\href{https://huggingface.co/sentence-transformers/all-MiniLM-L6-v2}{\texttt{all-MiniLM-L6-v2}}} model.

\begin{wraptable}{l}{0.44\textwidth}
\caption{The optimal number of hidden layers $H$ in the encoder of the teacher \textbf{T} for each dataset and different topic counts $K$.} \label{tab:HypOpt}
\begin{adjustbox}{width=\linewidth}
  \begin{tabular}{ | c  c  c || c  c  c | } \hline
    \textbf{Dataset} & \textbf{\textit{K}} & \textbf{\textit{H}} & \textbf{Dataset} & \textbf{\textit{K}} & \textbf{\textit{H}} \\ \hline \hline
    \multirow{4}{*}{\textbf{20NG}} & \multirow{4}{*}{\makecell{20 \\ 50 \\ 100}} & \multirow{4}{*}{\makecell{1 \\ 1 \\ 5}} & \multirow{4}{*}{\textbf{M10}} & 10 & 4 \\
    & & & & 20 & 5 \\
    & & & & 50 & 2 \\
    & & & & 100 & 3 \\ \hline
\end{tabular}
\end{adjustbox}
\end{wraptable}
Using the Bayesian optimization framework of OCTIS, we have calculated the optimal number of hidden layers $H$ in the teacher's encoder (which takes as input the concatenation of a document's contextualized and BoW representations) from the set $\{1,2,\hdots,10\}$ that maximizes the NPMI for the teacher. As shown in Table \ref{tab:HypOpt}, on 20NG dataset, we found $H=1$ for topic count $K \in \{20, 50\}$ and $H= 5$ for $K=100$;  on M10, we observed $H=4$ for $K=10$, $H=5$ for $K=20$, $H=2$ for $K=50$, and $H=3$ for $K=100$. Each hidden layer of the teacher contains 100 neurons. 

We have tuned the hyperparameters $\alpha \in [0,1]$ and $t \in \{1,2,\hdots,5\}$ for \textbf{SKD} in OCTIS. For performance analysis, we compare these models with \textbf{ProdLDA} \cite{srivastava2017autoencoding}, \textbf{NeuralLDA} \cite{srivastava2017autoencoding}, \textbf{Embedded Topic Model} (\textbf{ETM}) \cite{dieng-etal-2020-topic} and \textbf{LDA} \cite{blei2003latent}, already implemented in OCTIS.  We use the default parameters unless otherwise mentioned. All models are trained for 100 epochs with a batch size of 64. Each reported performance score is the median over 5 runs (except for \textbf{T}, where we use a single run as it must be frozen for KD). \\

\begin{figure}
    \centering
    \includegraphics[width=\textwidth]{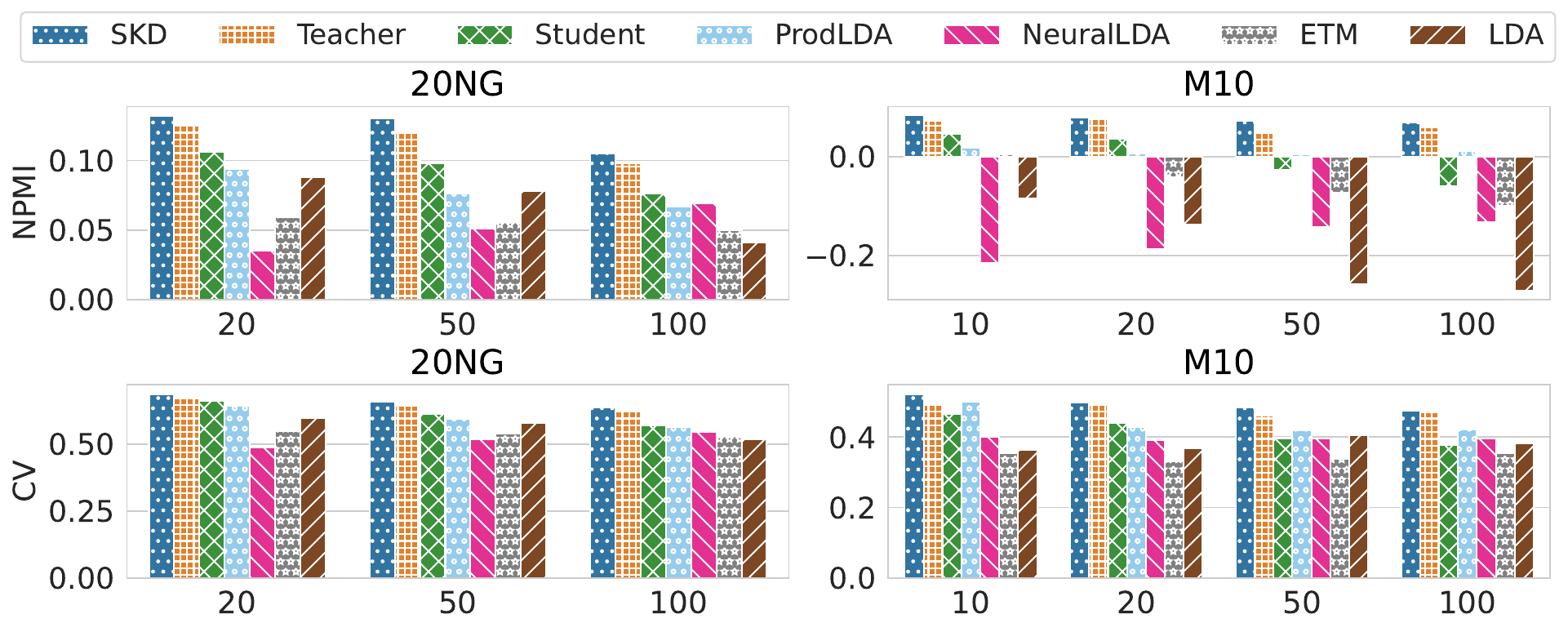}
    \caption{Coherence scores (\textbf{NPMI} and \textbf{CV}) for different topic models on two datasets: \textbf{20NG} and \textbf{M10}. The X-axis is marked with the topic counts used for each dataset.}
    \label{fig:Compare}
\end{figure}

\section{Results}
Models \textbf{S} and \textbf{SKD} contain the same number of parameters, which is smaller than that of \textbf{T}. The sizes of all the models depend on the SBERT dimension, the number and size of hidden layers, the number of topics, and the vocabulary size. For example, for 20 topics in 20NG, \textbf{T} takes 6.14 MB while \textbf{SKD} 2.74 MB (for parameters and buffers) -- a reduction in model size by $55.4\%$. In general, the compression ranged from $37.6\%$ to $56.3\%$. 

Fig. \ref{fig:Compare} shows the coherence scores for each topic model for all topic settings and datasets. \textbf{SKD} achieves the highest NPMI and CV scores. Among \textbf{T}, \textbf{S}, and \textbf{SKD}, we find \textbf{SKD} performs much better than \textbf{S} and even modestly better than \textbf{T}. On 20NG, the NPMI scores of (\textbf{T}, \textbf{S}, \textbf{SKD}) are $(0.125, 0.106, 0.132)$ for $K=20$, $(0.121, 0.098, 0.130)$ for $K=50$, and $(0.098, 0.076, 0.105)$ for $K=100$, so the maximum gain of \textbf{SKD} over \textbf{S} is $38.2\%$ and that over \textbf{T} is $7.4\%$. Similarly on M10, the NPMI scores are $(0.073, 0.046, 0.084)$ for $K=10$, $(0.076, 0.037, 0.08)$ for $K=20$, $(0.053, -0.027, 0.073)$ for $K=50$, and $(0.059, -0.06, 0.07)$ for $K=100$.
Thus, on M10, \textbf{SKD} improves NPMI of \textbf{S} by over $100\%$ for $K \in \{50, 100\}$, and that of  \textbf{T} by at most $37.7\%$. Student outperforming the teacher  is surprising but has been reported earlier for \textit{supervised} tasks \cite{furlanello2018born, zhang2019your}. 

When we deleted any one of the two loss terms from $\mathcal{L}_{\KD}$ in Eq. \eqref{eq:kdloss}, NPMI and CV of \textbf{SKD} dropped (see Table \ref{tab:ABLATION}). 
Thus, although the simpler model and weaker SBERT lower the student's performance, the knowledge distilled from the teacher's encoder and decoder vastly improves it. 
\begin{table}[!htbp]
\centering
\caption{Ablation study for the distillation loss term  defined in Eq. \eqref{eq:kdloss}. For each metric, the median over five independent runs for each topic count is mentioned.}
\label{tab:ABLATION}
\begin{adjustbox}{width=\linewidth}
  \begin{tabular}{ | c | c c c | c c c || c c c c | c c c c|} \hline
    \multirow{3}{*}{\textbf{KD-loss} ($\mathcal{L}_{\KD}$)} & \multicolumn{6}{c||}{\textbf{20NG}} & 
    \multicolumn{8}{c|}{\textbf{M10}} \\ \cline{2-15} 
    & \multicolumn{3}{c|}{\textbf{NPMI}} & \multicolumn{3}{c||}{\textbf{CV}} & \multicolumn{4}{c|}{\textbf{NPMI}} & \multicolumn{4}{c|}{\textbf{CV}}\\ \cline{2-15}
    & $\mathbf{20}$ & $\textbf{50}$ & $\mathbf{100}$ & $\mathbf{20}$ & $\textbf{50}$ & $\mathbf{100}$ & $\mathbf{10}$ & $\mathbf{20}$ & $\textbf{50}$ & $\mathbf{100}$ & $\mathbf{10}$ & $\mathbf{20}$ & $\textbf{50}$ & $\mathbf{100}$ \\\hline \hline
    $\mathcal{L}_{\text{KD-2W}} + \mathcal{L}_{\text{KD-CE}}$ & \textbf{0.132} & \textbf{0.130} & \textbf{0.105} & \textbf{0.687} & \textbf{0.657} & \textbf{0.638} & \textbf{0.084} & \textbf{0.080}  & \textbf{0.073} & \textbf{0.070} & \textbf{0.522} & \textbf{0.499} & \textbf{0.485} & \textbf{0.475} \\
    $\mathcal{L}_{\text{KD-2W}}$ & 0.109 & 0.114 & 0.089 & 0.659 & 0.638 & 0.615 & 0.051 & 0.049 & 0.037 & 0.043 & 0.498 & 0.479 & 0.459 & 0.452 \\
    $\mathcal{L}_{\text{KD-CE}}$ & 0.110 & 0.105 & 0.083 & 0.653 & 0.629 & 0.588 & 0.042 & 0.052 & 0.016 & 0.023 & 0.485 & 0.464 & 0.425 & 0.425 \\ \hline
  \end{tabular}
 \end{adjustbox}
\end{table}

The higher performance of the contextualized topic models over other topic models agrees with similar results in \cite{bianchi2021pre, bianchi2021cross}. 
In Table \ref{tab:manualEval}, we compare qualitatively some aligned topics learned by \textbf{T}, \textbf{S}, and \textbf{SKD} from the 20NG corpus. For the first three topics, \textbf{SKD} displays  more word overlap than \textbf{S} with the corresponding topics from \textbf{T}, showing that \textbf{T} and \textbf{SKD} learn similar topic-word distributions. Interestingly, the fourth topic  from \textbf{SKD} contains more healthcare-related words than the fourth topic from \textbf{T} although the latter is also primarily on healthcare; this shows that \textbf{SKD} can produce more coherent topics than \textbf{T}.

\begin{table*}[!htbp]
\centering
\caption{Some selected topics output when \textbf{T}, \textbf{S}, and \textbf{SKD} models are run on the 20NG corpus for 20 topics. If a word in a topic from \textbf{S} or \textbf{SKD} is shared with the corresponding topic in \textbf{T}, then it is in \textbf{bold} otherwise it is in \textit{italic}.} \label{tab:manualEval}
\begin{adjustbox}{width=.94\linewidth}
  \begin{tabular}{ | c | c | l | } \hline
    \textbf{Model} & {\textbf{ID}} & \multicolumn{1}{c|}{\textbf{Topics}} \\ \hline \hline
    \multirow{4}{*}{\textbf{T}} 
        & $0$ &  gun, law, firearm, crime, weapon, assault, amendment, state, police, permit \\
        & $11$ & russian, turkish, people, village, genocide, armenian, muslim, population, greek, army \\ 
        & $17$& oil, engine, ride, front, road, chain, bike, motorcycle, water, gas \\ 
        & $3$& health, make, president, patient, medical, people, doctor, disease, work, year \\ \hline \hline
    \multirow{4}{*}{\textbf{S}} 
        & $0$ & \textbf{law}, \textit{people}, \textbf{state}, \textit{government}, \textbf{gun}, \textbf{amendment}, \textit{constitution}, \textbf{firearm}, \textbf{crime}, \textit{privacy} \\
        & $1$ & \textbf{armenian}, \textbf{village}, \textit{soldier}, \textit{soviet}, \textbf{muslim}, \textit{troop}, \textbf{turkish}, \textbf{russian}, \textbf{genocide}, \textit{land}  \\
        & $17$ & \textbf{engine}, \textit{car}, \textit{mile}, \textbf{ride}, \textbf{bike}, \textbf{oil}, \textbf{front}, \textit{wheel}, \textbf{motorcycle}, \textit{tire} 
        \\ 
        & $7$ & \textbf{medical}, \textbf{disease}, \textit{study}, \textit{treatment}, \textbf{doctor}, \textbf{patient}, \textbf{health}, \textit{food}, \textit{risk}, \textit{percent} \\ \hline \hline
     \multirow{4}{*}{\textbf{SKD}} 
        & $0$ & \textbf{gun}, \textbf{law}, \textbf{weapon}, \textbf{firearm}, \textbf{amendment}, \textbf{crime}, \textit{bill}, \textit{assault}, \textit{constitution}, \textbf{police} \\ 
        & $11$ & \textbf{turkish}, \textbf{genocide}, \textbf{armenian}, \textbf{russian}, \textbf{village}, \textbf{population}, \textit{israeli}, \textit{war}, \textit{attack}, \textbf{muslim} \\ 
        & $17$ & \textbf{ride}, \textbf{engine}, \textit{car}, \textbf{bike}, \textbf{motorcycle}, \textbf{front}, \textbf{oil}, \textit{motor}, \textbf{road}, \textit{seat} \\ 
        & $3$ & \textbf{health}, \textbf{medical}, \textbf{doctor}, \textbf{disease}, \textbf{patient}, \textit{insurance}, \textit{treatment}, \textit{drug}, \textit{care}, \textit{risk}  \\ \hline

\end{tabular}
\end{adjustbox}
\end{table*}

\section{Conclusion}
We have proposed a 2-Wasserstein loss-based knowledge distillation framework to compress a contextualized topic model. Experiments on two datasets show that the pruned topic model produces topics with coherence better than that of the topics produced by the student and even the larger teacher model. This is a new method for neural topic distillation. In the future, we would like to study it analytically and apply it to distill knowledge across other neural topic models.

\end{document}